\documentclass{article} % For LaTeX2e
\usepackage{iclr2022_conference,times}

\usepackage{amsmath,amsfonts,bm}

\def\eqref#1{equation~\ref{#1}}
\def\1{\bm{1}}

\DeclareMathAlphabet{\mathsfit}{\encodingdefault}{\sfdefault}{m}{sl}
\SetMathAlphabet{\mathsfit}{bold}{\encodingdefault}{\sfdefault}{bx}{n}

\usepackage{hyperref}
\usepackage{url}
\usepackage{amsthm}
\usepackage{amssymb}
\usepackage{multirow}
\usepackage{algpseudocode, algorithm}
\usepackage{authblk}

\usepackage{tabularx}
\usepackage{tikz}

\usepackage{lipsum}

\usepackage{enumerate}
\usepackage{caption}
\usepackage{subcaption}

\DeclareMathOperator*{\mmax}{mellowmax}

\title{Count-Based Temperature Scheduling for Maximum Entropy Reinforcement Learning}

\author[1 \thanks{Correspondence to: dailinh@uci.edu}]{Dailin Hu}
\author[2]{\ Pieter Abbeel}
\author[1]{\ Roy Fox}
\affil[1]{Department of Computer Science, University of California, Irvine}
\affil[2]{Department of Electrical Engineering and Computer Science, \linebreak University of California, Berkeley}

\iclrfinalcopy % Uncomment for camera-ready version, but NOT for submission.
\begin{document}

\maketitle

\begin{abstract}
Maximum Entropy Reinforcement Learning (MaxEnt RL) algorithms such as Soft Q-Learning (SQL) and Soft Actor–Critic trade off reward and policy entropy, which has the potential to improve training stability and robustness. Most MaxEnt RL methods, however, use a constant tradeoff coefficient (temperature), contrary to the intuition that the temperature should be high early in training to avoid overfitting to noisy value estimates and decrease later in training as we increasingly trust high value estimates to truly lead to good rewards. Moreover, our confidence in value estimates is state-dependent, increasing every time we use more evidence to update an estimate. In this paper, we present a simple state-based temperature scheduling approach, and instantiate it for SQL as Count-Based Soft Q-Learning (CBSQL). We evaluate our approach on a toy domain as well as in several Atari 2600 domains and show promising results.
\end{abstract}

\section{Introduction}
Deep Reinforcement Learning (RL) methods that use neural-network function approximators to learn control policies have shown great performance in domains including games \citep{mnih2015human,silver2017mastering}, robotic control \citep{haarnoja2017reinforcement}, and autonomous driving~\citep{sallab2017deep}. The training of such a function approximator, however, is often very sensitive to hyperparameter tuning in different environments~\citep{henderson2018deep}. Recent work in Maximum Entropy Reinforcement Learning (MaxEnt RL)~\citep{fox2016taming,haarnoja2018soft} trades off maximizing the policy values with its entropy, producing policies that are more robust to disturbances in the environment dynamics and reward function~\citep{haarnoja2017reinforcement,eysenbach2021maximum}.

MaxEnt RL follows from the maximum entropy principle~\citep{jaynes2003probability}, which states that one should prefer maximally uninformed solutions, subject to the available evidence. MaxEnt RL therefore maximizes average policy entropy, subject to an attainable level of policy value. The Lagrangian of this constrained optimization problem trades off the traditional RL objective of maximum policy value with the expected policy entropy. The Lagrange multiplier is an \emph{inverse-temperature} parameter $\beta$ that determines the relative importance of the value and entropy terms, and corresponds to a level of value that we believe we can attain. Intuitively, in early stages of training, $\beta$ should be lower to introduce higher stochasticity as we have low confidence that the current value estimates are attainable. During training, we become increasingly more confident in our value estimates, and we increase $\beta$ such that the policy stochasticity decreases and eventually approaches zero. As $\beta \to \infty$, the conventional RL learning objective is recovered.

Most MaxEnt RL algorithms, such as Soft Q-Learning (SQL)~\citep{haarnoja2017reinforcement}, use a constant temperature throughout training. The log-partition function in SQL's soft Bellman backup operator allows it to put higher weight on the policy entropy when the temperature is high, or approximate reward maximization arbitrarily well as the temperature decreases to 0. Selecting a constant inverse-temperature $\beta$ may not reflect the changing confidence of the value estimates throughout learning. Even when there exists a constant $\beta$ that works well, it is domain-dependent, which makes it hard to tune as a hyperparameter. Soft Actor–Critic (SAC)~\citep{haarnoja2018soft} adjusts the temperature $\beta^{-1}$ automatically with stochastic gradient descent, but often has poor performance in discrete action spaces~\citep{christodoulou2019soft}. %Unfortunately, however, the instability in early stages of training often lead to inaccurate estimation of $\alpha$. 
\cite{fox2016taming} proposes a linear inverse-temperature schedule, in which $\beta_i = \kappa i$ in training step $i$. This corresponds to the intuition that $\beta$ should increase throughout learning, but the coefficient $\kappa$ remains a domain-dependent hyperparameter to be tuned. 
%(DH): More related paper on this...? Referencing Litian's work?

Importantly, our confidence in SQL's value estimates is very much state-dependent. Value estimates that have been updated more times, from more data, are more reliable. States that are more often experienced by early exploration will have their value estimates improve in accuracy, and their $\beta$ should increase. Later in learning, some of these states may not be encountered as often, becoming less important to train, and their $\beta$ should not increase as fast. When exploration reaches novel states, whose value estimates are less reliable, their $\beta$ should start very low again, even late in training.

% contents of this paper
In this paper, we describe a simple state-dependent temperature scheduling method for MaxEnt RL based on a pseudo-count of state value updates derived from a CTS density model~\citep{ostrovski2017count}. As a concrete instance of this method, adapted to the SQL algorithm, we present the Count-Based Soft Q-Learning (CBSQL) algorithm. Evaluating CBSQL on 6 popular games on the Atari 2600 platform suggests that our scheduling method improves over DQN and SQL with a fixed temperature, and display the potential to reach state-of-the-art performance with Rainbow integration~\citep{hessel2018rainbow}.

\section{Preliminaries}
We address MaxEnt RL in discrete action spaces. In this section, we will introduce the notation and framework of MaxEnt RL. 
\subsection{Reinforcement Learning}
% MDP, Max Entropy RL
%The MaxEnt RL problem, same as any reinforcement learning problems, can be 
We consider environments modeled as a Markov Decision Process (MDP). We describe this process by a tuple $\langle\mathcal{S},\mathcal{A}, p, r\rangle$, where $\mathcal{S}$ represents a state space and $\mathcal{A}$ represents a discrete action space. $p(s_{t+1}| s_t, a_t): \mathcal{S} \times \mathcal{A} \rightarrow \Delta(\mathcal{S})$ represents the probability distribution of the next state $s_{t+1}$ given the current state $s_t$ and action $a_t$. $r_t = r(s_t, a_t): \mathcal{S}\times \mathcal{A} \rightarrow \mathbb{R}$ describes the reward for each transition $t$. The discounted return $R$, for a discount factor $0 \le \gamma < 1$, is defined as $R = \sum_t \gamma^t r_t$.

A reinforcement learning agent learns a policy $\pi(a_t | s_t)$ for interacting with its environment. The agent and the environment jointly induce a distribution $p_\pi(\xi)$ over trajectories $\xi = s_0, a_0, s_1, a_1, \ldots$. It is also convenient to define a state distribution $p^\gamma_\pi(s) = (1 - \gamma) \sum_t \gamma^t p_\pi(s_t = s)$, describing the distribution of $s_t$ at a time $t$ distributed geometrically with parameter $1 - \gamma$. This satisfies $\mathbb{E}_{\xi \sim p_\pi}[R(\xi)] = \tfrac1{1-\gamma} \mathbb{E}_{s \sim p^\gamma_\pi}[\mathbb{E}_{(a | s) \sim \pi}[r(s, a)]]$.

\subsection{Maximum Entropy Reinforcement Learning}
The MaxEnt RL objective augments the standard reinforcement learning objective of maximizing expected discounted rewards by adding an entropy term
\begin{equation}
    \pi^* = \arg \max_\pi \mathbb{E}_{s \sim p^\gamma_\pi}\left[ \mathbb{E}_{(a | s) \sim \pi}[ r(s, a) ] + \tfrac{1}{\beta} \mathbb{H}[\pi(\cdot | s)]\right], \label{eq:opt}
\end{equation}
where $\beta$ is an \emph{inverse-temperature} parameter that controls the stochasticity of the optimal policy by determining the relative importance between the reward and policy entropy $\mathbb{H}[\pi]$. In early stages of training, $\beta$ should intuitively be assigned a small value to induce stochasticity, and in later stages of training $\beta \rightarrow \infty$ to approach the deterministic behavior that optimizes the standard reinforcement learning objective. 

\subsection{Soft Q-Learning (SQL) and the Mellowmax Operator}
RL methods often involve the state–action value function $Q(s, a) = \mathbb{E}[R | s_0 = s, a_0 = a]$ and the Bellman operator $\mathcal{B}[Q](s, a) = r(s, a) + \gamma \mathbb{E}_{(s' | s, a) \sim p}[\max_{a'} Q(s', a')]$.
The MaxEnt RL objective suggests a \emph{soft} Bellman operator~\citep{rubin2012trading}
%Based on the MaxEnt RL framework in section 2.2, \cite{haarnoja2017reinforcement} proposed the fixed point iteration for continuous domains. The similar iteration on discrete domains can be described as
\begin{align}
    \mathcal{B}[Q](s, a) &= r(s, a) + \gamma \mathbb{E}_{(s' | s, a) \sim p}\left[\max_\pi \left( \mathbb{E}_{(a' | s') \sim \pi}[ Q(s', a') ] + \tfrac{1}{\beta} \mathbb{H}[\pi(\cdot | s')] \right) \right] \label{eq:bellman1} \\
    &= r(s, a) + \gamma \mathbb{E}_{(s' | s, a) \sim p}\left[\tfrac{1}{\beta} \log \sum_{a' \in \mathcal{A}} \exp(\beta Q(s', a'))\right] \qquad \forall s \in \mathcal{S}, a \in \mathcal{A}, \label{eq:bellman}
\end{align}
% \begin{equation}
%     V(s_t) \leftarrow \frac{1}{\beta} \log \sum_{a \in \mathcal{A}} \frac{1}{|\mathcal{A}|} \exp(\frac{1}{\beta} Q(s_t, a)).
% \end{equation}
where the optimizer in (\ref{eq:bellman1}) is the softmax policy $\pi(a | s) \propto \exp(\beta Q(s, a))$, and the log-sum-exp expression in (\ref{eq:bellman}) is the log-partition function, also called the mellowmax operator~\citep{asadi2017alternative}.
% 
% This fixed-point iteration differs from the classic Q-value iteration in that it centers around the mellowmax operator\citep{asadi2017alternative}:
% 
% \begin{equation}
%     \textit{mm}_\beta (x) = \frac{1}{\beta}\log\frac{1}{n} \sum_{i=1}^n \exp(\beta x_i).
% \end{equation}.
% 
Mellowmax is non-decreasing in $\beta$~\citep{kim2019deepmellow} and a non-expansion for a fixed $\beta$ under the supremum norm~\citep{fox2016taming}, and as $\beta \to \infty$ it converges to pure maximization.

Soft Q-Learning~\citep{fox2016taming,haarnoja2017reinforcement} uses a model-free empirical estimate of the soft Bellman operator as the target for learning a Q function. When the Q function has a tabular representation, the contraction property of the soft Bellman operator guarantees convergence to the operator's fixed point, the softmax of which is the optimal policy in (\ref{eq:opt})~\citep{fox2016taming}.

\section{Count-Based Temperature Scheduling}

\subsection{Motivation}

Empirical evidence from SQL suggests that a state-independent linear temperature scheduling, i.e. using $\beta_i = \kappa i$ in iteration $i$, can achieve good performance~\citep{fox2016taming, grau2018soft}. \cite{fox2019toward} proposes a closed-form state-dependent expression for the temperature that completely eliminates bias in entropy-regularized value updates in two-action environments. If the gap between the action values $Q(s, a=0) - Q(s, a=1)$ has Gaussian uncertainty with mean $\mu_A$ and variance $\sigma^2_A$, the prescribed inverse-temperature is $\beta(s) = \frac{2 \mu_A}{\sigma^2_A}$. The Gaussian assumption is motivated by the central limit theorem, which also suggests that the asymptotic behavior of $\beta$ is to grow linearly in the number of data points used to estimate the value gap at each state.

We propose Count-Based Soft Q-Learning (CBSQL), an algorithm based on SQL that uses a state-dependent temperature schedule in which $\beta(s)$ grows linearly with the number of times that the algorithm updates $Q(s, a)$, for any action $a$. Formally, let $n(s, a)$ be the count of sampled data points of the form $(s, a, r, s')$ used thus far in value updates. Then the inverse-temperature in CBSQL is $\beta(s) = \kappa \sum_a n(s, a)$, with $\kappa > 0$ a constant hyperparameter.

In addition to the aforementioned empirical and theoretical evidence supporting linear count-based scheduling, more insight can be gained by comparing two families of successful RL algorithms. The first family, consisting of such algorithms as G-Learning~\citep{fox2016taming}, SQL~\citep{haarnoja2017reinforcement}, Path Consistency Learning (PCL)~\citep{nachum2017bridging}, and Soft Actor–Critic (SAC)~\citep{haarnoja2018soft}, aims to learn a policy that is the softmax of a value function $Q(s, a)$ with a low temperature. In iteration $i$, the policy target is
%Suppose we have a fixed critic function $Q(s,a)$. In MaxEnt RL algorithms such as G-learning, SQL and SAC, the $k_{th}$ policy update can be generalized as 
\begin{equation}
    \pi_i (a| s) \propto \pi_0(a|s) \exp \beta_i(s) Q_i(s, a) \qquad \forall s\in \mathcal{S}, a \in \mathcal{A}, \label{eq:type1}
\end{equation}
with $\pi_0$ the uniform policy, and the policy is either updated toward the target (\ref{eq:type1}), or set equal to it. Note that in their original formulations, most of these algorithms use a constant state-independent $\beta$. The second family of algorithms, consisting of Relative Entropy Policy Search (REPS) \citep{peters2010relative}, $\Psi$-learning~\citep{rawlik2010approximate}, Trust Region Policy Optimization (TRPO)~\citep{schulman2015trust}, and Trust-PCL~\citep{nachum2017trust}, aims to use value estimates to gradually update the softmax policy. Instead of the policy entropy, which is the Kullback–Leibler (KL) divergence from a uniform prior policy, these algorithms consider a KL term with the current policy as the prior for the update. Moreover, instead of a low temperature $\beta^{-1}$, these algorithms place a large coefficient $\kappa^{-1}$ on the KL term, to induce small updates in a “trust region”. In iteration $i$, the policy update is therefore
\begin{equation}
    \pi_i(a|s) \propto \pi_{i-1}(a|s) \exp \kappa_i(s) Q_i(s, a)\propto \pi_0(a|s)\exp\left(\sum_{j\le i} \kappa_j(s) Q_j(s,a)\right). \label{eq:type2} %\qquad \forall s\in \mathcal{S}, a \in \mathcal{A}. 
\end{equation}  
The two types of learning processes (\ref{eq:type1}) and (\ref{eq:type2}) can have different properties, because $Q_i$ can change significantly between iterations. For the sake of our intuitive argument here, imagine that $Q$ could be approximated by a constant, and that the inverse-temperatures $\kappa_i(s)$ used in trust-region algorithms were also a constant $\kappa$.
Then combining the above two equations, we have
\begin{equation}
    \beta_i(s) \approx \sum_{j\le i} \kappa_j(s) \approx \kappa i,
\end{equation}
where $\kappa$ is a small constant, and $i$ is the number of times that the update equation (\ref{eq:type2}) is applied in state $s$. In practice, updates are not applied to all actions in state $s$ at the same time, leading to the heuristic definition $i = \sum_a n(s, a)$. %We adapt this temperature scheduling to SQL (See appendix).

% The major difference between our method and G-learning, which schedules $\beta$ proportional to the total training timesteps, is that we see $\beta_k$ as a function of the states. In other words, $\beta$ is proportional to the number of times the critic has updated state $s$. This scheduling method combines two novel but very different policy update approaches, and corresponds to what we expect of the temperature. The justification for our scheduling method is relatively loose. In actual training, the critic function will not be fixed, and $\eta$ of course is not always the same value.

\subsection{Pseudo-counts}

% (discuss large state spaces, how we can't count but also don't want to because updates in similar states should also count to some extent, how this is solved in count-based exploration, how we update the density models in CBSQL)

Directly recording the number of times that the value update has been applied to state $s$ is not useful in practical settings, where the state space is large, and most states will rarely be visited more than once. Moreover, a tabular mapping from $s$ to $i$ would fail to capture the similarity between different states that a Q function approximator does leverage, and would vastly underestimate the effective number of times that $Q(s, a)$ has been updated in states similar to $s$.

Instead, we use a pseudo-count method derived by
\cite{DBLP:journals/corr/BellemareSOSSM16} from a simplified pixel-level CTS density model~\citep{bellemare2014skip}. Define $\rho$ to be the CTS density model on the state space $\mathcal{S}$. Let $\rho_k (s)$ be the probability assigned by the model to $s \in \mathcal{S}$ after $k$ updates of the model. Let $\rho'_k(s)$ be the probability that the model would assign to $s$ if it were updated on $s$ one more time. The pseudo-count can then be defined as
\begin{equation}
    n_k(s) = \frac{\rho_k(s)(1-\rho'_k(s))}{\rho'_k(s) - \rho_k(s)}. \label{eq:pseudo}
\end{equation}
Note that $k$ is different from $i$ in the last section and represents the total number of updates in all states. The effective number of times that state $s$ has been updated is its pseudo-count $n_k(s)$, suggesting the state-dependent inverse-temperature
\begin{equation}
    \beta(s) = \kappa \cdot n_k(s),
\end{equation}
where $\kappa$ is a small constant. Throughout our experiments, we set $\kappa = 0.01$.

% \subsection{Mellowmax with Dynamic Temperature}
% In this section we discuss some interesting properties of the mellowmax operator, and present proof for the convergence of SQL with dynamic temperature update. 

% \cite{fox2016taming} proved that the mellowmax operator with a fixed temperature is a contraction and converge to a fixed point. This result does not directly apply to a dynamic temperature.
In summary, we propose a model-free reinforcement learning algorithm that, on experience $(s, a, r, s')$, uses the SQL update rule
% \begin{equation}
%     Q(s_t, a_t) \leftarrow r_t + \gamma \mathbb{E}_{s_{t+1} \sim P}[V(s_{t+1})],
% \end{equation}
% \begin{equation}
%     V(s_t) \leftarrow \frac{1}{\beta(s_t)} \log \frac{1}{|\mathcal{A}|}\sum_{a\in\mathcal{A}} \exp(\beta(s_t)Q(s_t, a)),
% \end{equation}
% \begin{equation}
%     \beta(s) \leftarrow \beta(s) + 1.
% \end{equation}
\begin{equation}
    Q(s, a) \gets r + \gamma \mmax_{a';\beta(s')} Q(s', a') = r + \tfrac{\gamma}{\beta(s')} \log \tfrac{1}{|\mathcal{A}|}\sum_{a'\in\mathcal{A}} \exp(\beta(s')Q(s', a')).
\end{equation}
In tabular experiments, we also update
\begin{equation}
    n(s) \gets n(s) + 1,
\end{equation}
whereas in deep-learning experiments, we update the density model with state $s$, and consider the pseudo-count (\ref{eq:pseudo}).
The inverse-temperature is then scheduled as $\beta(s) = \kappa \cdot n(s)$.
 We present the pseudocode in Alg 1.

\begin{algorithm}[]
\caption{Count-Based Soft Q Learning (SQL)}
\begin{algorithmic}
\State Initialize Q network parameters $\theta$ 
\State Initialize target Q network $\overline{\theta} \gets \theta$
\State Initialize an empty replay buffer $\mathcal{D} \gets \emptyset$ 
\State Initialize a density model $\rho$
% \State Initialize $k$ \gets 1

\For{each iteration}
    \For {each step $t$ in the rollout}
        \State In state $s_t$, sample action $a_t$ from the $\epsilon$-greedy policy for $Q_\theta(s_t, \cdot)$%, i.e. $a_t = \arg\max_a Q_\theta(s_t, a)$ with probability $1 - \epsilon$, uniform otherwise
        \State Execute action $a_t$ and observe reward $r_t$ and new state $s_{t+1}$
        \State Store the transition $(s_t, a_t, r_t, s_{t+1})$ into the replay buffer $\mathcal{D}$
    \EndFor
    \For {each gradient step}
        \State Sample random batch $(s, a, r, s')$ from $\mathcal{D}$
        % \State Update CTS model $\rho$
        \State $\beta \gets \kappa \cdot \frac{\rho(s')(1-\rho'(s'))}{\rho'(s') - \rho(s')}$
        \State $y = r + \gamma \mmax_{a';\beta} Q_{\bar{\theta}}(s', a')$
        \State Perform gradient descent on $(y - Q_\theta (s, a))^2$
        \State Update the density model with state $s$
        \State Every \texttt{target\_freq} steps, update $\bar{\theta} \gets \theta$
    \EndFor
\EndFor

\end{algorithmic}
\end{algorithm}

\section{Experiments}

In this section, we evaluate our approach in tabular representation on a toy domain, and in deep learning on six popular Atari 2600 environments.
\subsection{Tabular Representation}
We compare our approach to DQN and SQL on a toy domain of noisy transitions in a chain-linked graph. This environment is small enough to directly find the optimal policy.

\begin{figure}[H]
     \centering
     \begin{subfigure}[b]{0.49\textwidth}
         \centering
         \includegraphics[width=\textwidth]{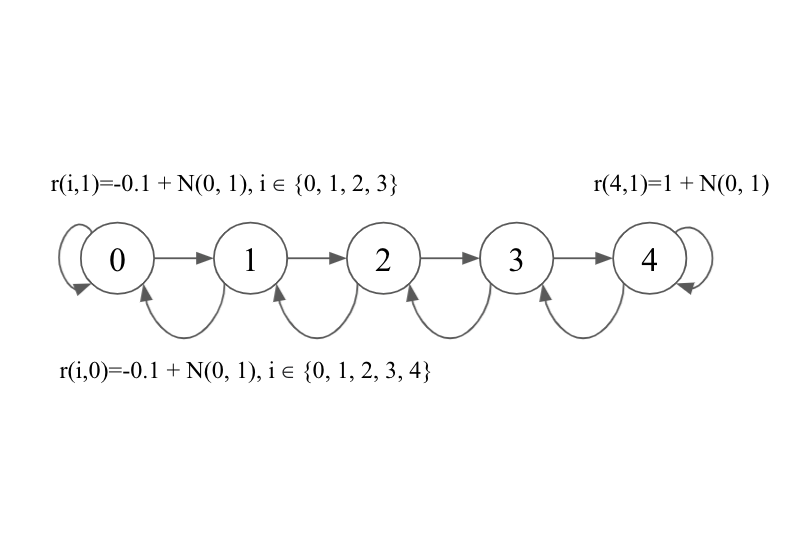}
         \caption{Domain Demonstration}
         \label{fig:chain}
     \end{subfigure}
     \hfill
     \begin{subfigure}[b]{0.49\textwidth}
         \centering
         \includegraphics[width=\textwidth]{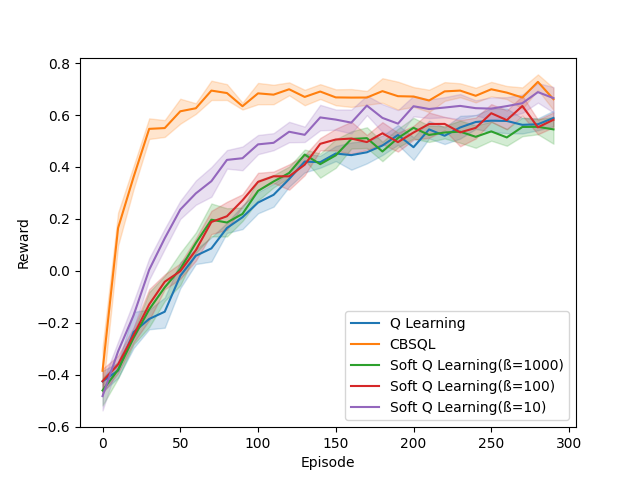}
         \caption{Rewards averaged over 10000 runs}
         \label{fig:chain_res}
     \end{subfigure}
     \hfill
    \caption{Noisy Chain-Walk Problem}
    \label{fig:three graphs}
\end{figure}

We define a noisy chain-walk problem with five states, each state with two possible actions (Figure \ref{fig:chain}). The agent always starts at state 0 and each episode ends after 5 steps. The agent receives a reward of $+1$ when it takes action 1 at state 4, and a reward of $-0.1$ when it takes any action at any other states. The reward that agent receives is always corrupted with an additive Gaussian noise of $\mathcal{N}(0, 1)$. %See Fig 1(a) for more information.

The optimal expected reward is $0.6$. %, and the actual reward is $0.6 + N(0, 1)$. 
Figure \ref{fig:chain_res} shows the rewards averaged over 1000 runs comparing Q-learning, SQL with different constant inverse-temperature parameters $\beta \in \{10, 100, 1000\}$, and CBSQL with a tabular representation. %For simplicity, instead of using CTS to produce a pseudo-count, we keep a dictionary for the number of times CBSQL agent has visited to each state. 
These results suggest that CBSQL can converge significantly faster and obtain higher rewards than both SQL and Q learning within 300 episodes of training in this simple domain. 
% The noisy binary tree problem has three states and each state has two actions. The agent starts and state 0 to go to either state 1 or 2 with 0 reward, and from state 1 or 2 taking different actions yield different rewards. The reward agent receives is corrupted with additive Gaussian noise $N(0, 1)$. See Fig. 1(a) for more information. 

\subsection{Neural Network Representation}
\begin{figure}[H]
    \centering
    \includegraphics[width=\linewidth]{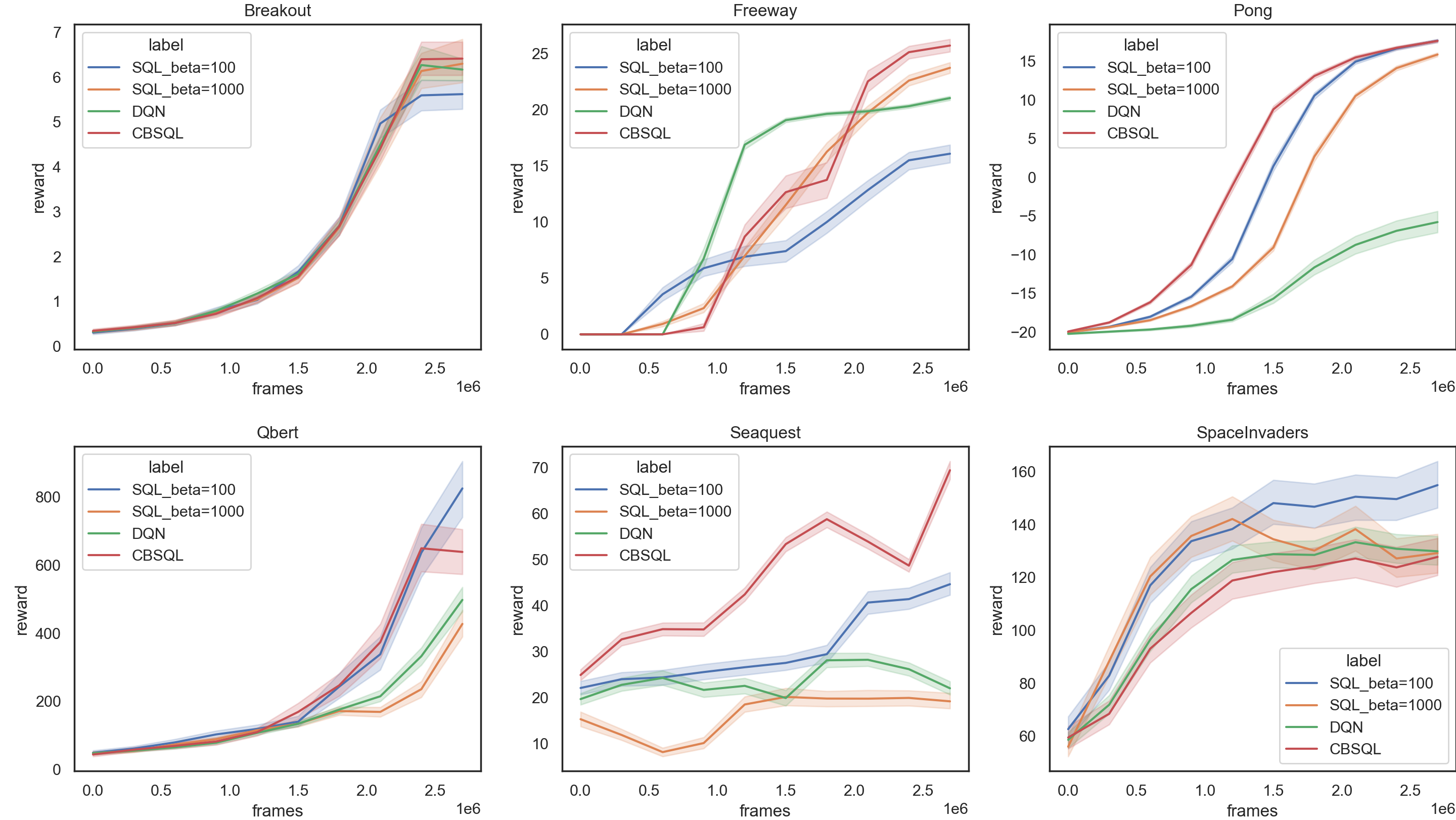}
    \caption{CBSQL results compared with DQN and fixed-temperature SQL, without Rainbow. Rewards are averaged over 3 runs.}
    \label{fig:1}
\end{figure}

We evaluate our method on six Atari 2600 game from the arcade learning environment~\citep{bellemare2013arcade} and compare it with DQN and SQL under the same hyperparameter setting from \cite{mnih2015human}. We train the agents on 3M frames and record the rewards. CBSQL appears to outperform DQN and fixed-temperature SQL in 4 of the 6 games (Table \ref{table:1}, Figure \ref{fig:1}).

\begin{table}[h!]
\begin{center}
\small
\begin{tabular}[c]{ l|c c c c } 

\hline
Game & DQN & SQL($\beta=100$) & SQL($\beta=1000$) & CBSQL  \\
\hline\hline
Breakout & 5.9 ($\pm$5.9) & 5.9 ($\pm$4.5) & 5.1 ($\pm$4.7) & \textbf{8.2} ($\pm$6.1) \\
Freeway & 21.0 ($\pm$1.5) & 14.6 ($\pm$8.5) & 22.56 ($\pm$4.7) & \textbf{25.82} ($\pm$4.9) \\ 
Pong & 1.93 ($\pm$2.6) & \textbf{17.83} ($\pm$2.2) & 16.31 ($\pm$2.7) & 17.56 ($\pm$2.0) \\ 
Q*bert & 568.4 ($\pm$1101.9) & 828 ($\pm$1411.7) & 564.5 ($\pm$1097.5) & \textbf{875.3} ($\pm$1254.6) \\
Seaquest & 13.5 ($\pm$24.1) & 4 ($\pm$60.2) & 17.2 ($\pm$24.0) & \textbf{84.6} ($\pm$60.2) \\ 
Space Invaders & 132.7 ($\pm$113.2) &\textbf{158.9} ($\pm$128.5) & 132.25 ($\pm$118.4) & 138.9 ($\pm$112.8) \\
\hline

\end{tabular}

\caption{DQN, fixed-temperature SQL, and CBSQL average rewards ($\pm$ standard deviation). Raw scores are averaged over the last 100 testing episodes across 3 runs.} \label{table:1}

% \caption{Raw scores across games, averaged over last 100 episodes of 3 runs, each run 3M timesteps.} \\
\end{center}
\end{table}

DQN by itself is a powerful algorithm, but in recent years more extensions to it have been proposed that greatly improve its performance. Interestingly, SQL and CBSQL can be combined with many of these extensions, allowing us to compare these methods after integrating popular DQN extensions. We integrate CBSQL with Rainbow DQN~\citep{hessel2018rainbow}, a state-of-the-art reinforcement learning algorithm for memoryless agents, which includes multi-step targets~\citep{sutton2018reinforcement}, double Q-learning~\citep{hasselt2010double}, prioritized experience replay~\citep{schaul2015prioritized}, dueling networks~\citep{wang2016dueling}, distributional RL~\citep{DBLP:journals/corr/BellemareDM17}, and noisy networks~\citep{fortunato2017noisy}. %and achieves state-of-the-art performance on Atari 2600. 
All of these methods can be straightforwardly applied to soft Q-learning, with the exception of multi-step targets and distributional RL, which we discuss next.

\textbf{Multi-step learning.} Multi-step targets with a well-tuned number of steps $n$ can lead to faster learning in on-policy RL algorithms~\citep{sutton2018reinforcement} by trading off the bias and variance of the return estimates~\citep{kearns2000bias}. The n-step truncated return at time $t$ is $r^{(n)}_t = \sum_{k=0}^{n-1} \gamma^k r(s_{t+k}, a_{t+k})$. \cite{hessel2018rainbow} defined a multi-step variant of DQN by minimizing the alternative loss
$
    (r_t^{(n)} + \gamma^n\max_{a\in \mathcal{A}}Q_{\bar{\theta}}(s_{t+n}, a) - Q_\theta (s_t, a_t))^2
$
and demonstrated through empirical experiments that n-step targets can outperform single-step targets in DQN, despite the off-policy experience providing biased estimates of the n-step return. In SQL with inverse-temperature $\beta$, the n-step truncated return is
\begin{equation}
    \tilde{r}^{(n)}_t = r^{(n)}_t + \tfrac1\beta \sum_{k=1}^{n-1} \gamma^k \mathbb{H}[\pi(\cdot | s_{t+k})]. \label{eq:nstep}
\end{equation}
Unfortunately, empirical policy entropy estimates are often very noisy. It is also unclear which $\beta$ should be used in off-policy estimates of (\ref{eq:nstep}).
These considerations call for further study, and in this work we simply use 1-step returns for SQL and CBSQL.
% and in this work we simply estimate $\tilde{r}_t^{(n)}$ using $r^{(n)}_t$, and minimize
% \begin{equation}
%     \left(r_t^{(n)} + \tfrac{\gamma^n}{\beta(s_{t+n})}\log \tfrac{1}{|\mathcal{A}|}\sum_{a\in\mathcal{A}} \exp(\beta(s_{t+n}) Q_{\bar{\theta}} (s_{t+n}, a)) - Q_\theta (s_t, a_t)\right)^2.
% \end{equation}

\textbf{Distributional RL.} Unlike conventional RL which estimates the expected return, distributional RL estimates the distribution of the return at time $t$ over the stochasticity of the environment and the policy. The estimator uses a fixed $N$-dimensional vector $\bm{z}$ of values spaced evenly along the range $[v_{\min}, v_{\max}]$ of possible returns. The distribution of $Q(s_t, a_t)$ is then represented by a categorical distribution $\bm{p}_\theta (s_t, a_t)$ over the values of $\bm{z}$. Distributional DQN updates $\bm{p}_\theta(s, a)$ by minimizing its KL-divergence from a projected target distribution induced by the categorical distribution $\bm{p}_{\bar{\theta}}(s', a^*)$ over the values $r + \gamma \bm{z}$. Here the action $a^*$ is chosen greedily with $a^* = \arg\max_{a'} \bm{z}^\intercal \bm{p}_{\bar{\theta}}(s', a')$. We adapt distributional RL to soft Q-learning and CBSQL by defining a policy distribution of 
$$\pi(a' | s') = \frac{\exp \beta(s') \bm{z}^\intercal \bm{p}_{\bar{\theta}}(s', a')}{\sum_{\bar{a}'} \exp \beta(s') \bm{z}^\intercal \bm{p}_{\bar{\theta}}(s', \bar{a}')}$$

over the values $r + \gamma\left(\bm{z} -\tfrac1\beta \mathbb{D}[\pi(\cdot | s') \| \pi_0] \right)$. 

\begin{figure}[H]
     \centering
     \begin{subfigure}[b]{0.49\textwidth}
         \centering
         \includegraphics[width=\textwidth]{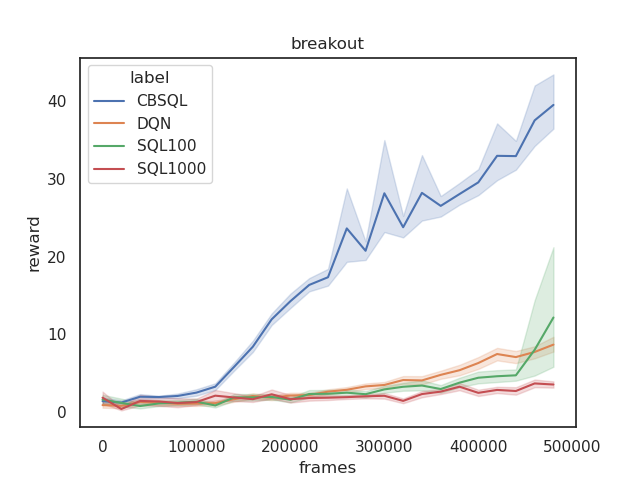}

     \end{subfigure}
     \hfill
     \begin{subfigure}[b]{0.49\textwidth}
         \centering
         \includegraphics[width=\textwidth]{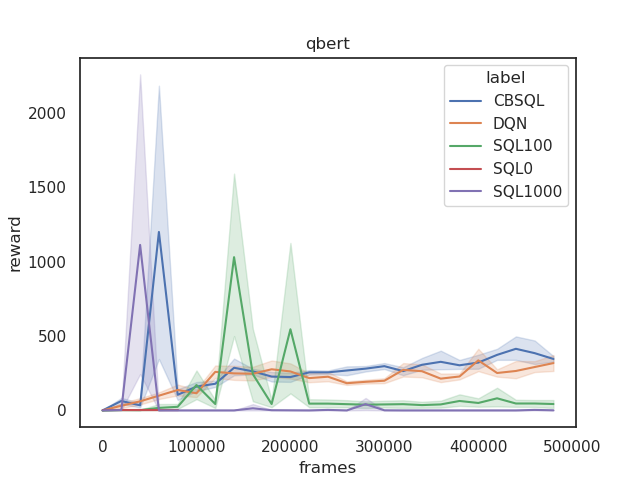}

     \end{subfigure}
     \hfill
    \caption{CBSQL results compared with DQN and fixed-temperature SQL, with Rainbow. Rewards are averaged over 5 runs.}
    \label{fig:Rainbow}
\end{figure}

We train the Rainbow-variations of SQL and CBSQL as described above over 500K frames each for the atari game Breakout and Q*bert against Rainbow-DQN(See Fig \ref{fig:Rainbow}). Rainbow-CBSQL out-performs Rainbow DQN in both these environments.

% \begin{figure}[H]
%     \begin{center}
%     \includegraphics[width=0.5\linewidth]{Rainbow_breakout.png}
%     \end{center}
%     \caption{}
%     \label{fig:2}
% \end{figure}
%  We train all the agents with Rainbow extensions over 1M frames and record the rewards. Rainbow-CBSQL outperforms Rainbow-DQN and Rainbow-SQL with constant temperatures in 3 out of 6 games and shows better performance than Rainbow-SQL in all 6 games. 

\begin{table}[h!]
\begin{center}
\small

\begin{tabular}{ l|cccc} 
\hline
Game & DQN & SQL($\beta=100$) & SQL($\beta=1000$) & CBSQL  \\
\hline\hline
Breakout & 10.2  ($\pm$ 4.9) & 5.5 ($\pm$ 3.6) & 3.6 ($\pm$ 2.1) & \textbf{39.9} ($\pm$ 15.4) \\
Q*bert & 356.0 ($\pm$ 202.8 ) & 30.0 ($\pm$ 60.0) & 1.0 ($\pm$ 4.9) & \textbf{370.0} ($\pm$ 43.0) \\
\hline

\end{tabular}
\end{center}
\caption{CBSQL results compared with DQN and SQL with fixed temperatures, combined with Rainbow. Raw scores averaged over last 50 testing episodes across 5 runs.} \label{table:2}
\end{table}

\section{Conclusion}

In this paper, we presented a simple method for temperature scheduling in soft Q learning which could be potentially applied to other maximum entropy reinforcement learning algorithms. %We discussed properties of the log-partition function component in SQL and proved that our algorithm converges. 
We showed through empirical experiments that our method can outperform DQN and SQL with or without the rainbow framework.

\bibliography{iclr2022_conference}
\bibliographystyle{iclr2022_conference}

\newpage
\appendix
\section{Hyperparameters}
\begin{table}[H]
    \begin{center}
    \begin{tabular}{l|c}
        Hyper-parameter & value \\
        \hline
         Discount factor $\gamma$& 0.99 \\
         Exploration $\epsilon$ & 0.01 \\
         Learning rate & 1 \\
    \end{tabular}
    \caption{Hyper-parameters for tabular experiments.}
    \label{tab:tabularcbsqlparam}    
    \end{center}
\end{table}

\begin{table}[H]
    \begin{center}
    \begin{tabular}{l|c}
        Hyper-parameter & value \\
        \hline
        Grey-scaling & True \\
        Observation down-sampling & (84, 84) \\
        Frames stacked & 4 \\
        Reward clipping & [-1, 1] \\        
         Discount factor $\gamma$ & 0.99 \\
         Initial exploration & 1 \\
         Final exploration & 0.1 \\
         Final exploration frame & 1000000 \\
         Learning rate & 0.00025 \\
         Replay buffer size & 1000000 \\
         Minibatch size & 32 \\
        Q network channels & [32, 64, 64]\\
        Q network filter size & $8 \times 8, 4 \times 4, 3 \times 3$ \\
        Q network stride & 4, 2, 1 \\
        Q network hidden units & 512 \\         
    \end{tabular}
    \caption{Hyper-parameters for DQN, SQL and CBSQL on Atari 2600. The values of all the hyper-parameters are based on the work from  \cite{mnih2015human}.}
    \label{tab:tabularcbsqlparam2}    
    \end{center}
\end{table}

\begin{table}[H]
    \begin{center}
    \begin{tabular}{l|c}
        Hyper-parameter & value \\
        \hline
        Grey-scaling & True \\
        Observation down-sampling & (84, 84) \\
        Frames stacked & 4 \\
        Reward clipping & [-1, 1] \\
        Discount factor $\gamma$ & 0.99 \\
        Learning rate & 0.0000625 \\
        Replay buffer size & 1000000 \\
        Minibatch size & 32 \\
        Q network channels & [32, 32, 64]\\
        Q network filter size & $8 \times 8, 4 \times 4, 3 \times 3$ \\
        Q network stride & 4, 2, 1 \\
        Q network hidden units & 512 \\
        Noisy net $\sigma_0$ & 0.5 \\
        Multi-step returns $n$ & 4 for DQN, 1 for SQL and CBSQL\\
        Distributional atoms & 51 \\
        Distributional min/max values & [-10, 10] \\
    \end{tabular}
    \caption{Hyper-parameters for DQN, SQL and CBSQL with Rainbow Integration on Atari 2600. The values of all the hyper-parameters are based on the work from \cite{hessel2018rainbow}. }
    \label{tab:tabularcbsqlparam3}    
    \end{center}
\end{table}

% \[
% \frac1\beta \log \sum_{a'} \pi_0(a' | s') \exp \beta Q(s', a') = \max_{\pi(\cdot | s')} \sum_{a'} \pi(a' | s') \beta Q(s', a') - D[ \pi(\cdot | s') \| \pi_0(\cdot | s') ]
% \]
% \[
% \sum_{a'} \pi(a' | s') \beta Q_{\bar{\theta}}(s', a') - D[ \pi(\cdot | s') \| \pi_0(\cdot | s') ] \qquad \pi = \arg\max_{\pi(\cdot | s')} \sum_{a'} \pi(a' | s') \beta Q_\theta(s', a') - D[ \pi(\cdot | s') \| \pi_0(\cdot | s') ]
% \]

\end{document}